# A Bayesian Approach to Constraint Based Causal Inference

Tom Claassen and Tom Heskes
Institute for Computer and Information Science
Radboud University Nijmegen
The Netherlands

## Abstract

We target the problem of accuracy and robustness in causal inference from finite data sets. Some state-of-the-art algorithms produce clear output complete with solid theoretical guarantees but are susceptible to propagating erroneous decisions, while others are very adept at handling and representing uncertainty, but need to rely on undesirable assumptions. Our aim is to combine the inherent robustness of the Bayesian approach with the theoretical strength and clarity of constraint-based methods. We use a Bayesian score to obtain probability estimates on the input statements used in a constraint-based procedure. These are subsequently processed in decreasing order of reliability, letting more reliable decisions take precedence in case of conflicts, until a single output model is obtained. Tests show that a basic implementation of the resulting Bayesian Constraint-based Causal Discovery (BCCD) algorithm already outperforms established procedures such as FCI and Conservative PC. It can also indicate which causal decisions in the output have high reliability and which do not.

## 1 Introduction: Robust Causal Discovery

In many real-world systems the relations and interactions between variables can be modeled in the form of a causal directed acyclic graph (DAG) $\mathcal{G}_C$ over a set of variables $\mathbf{V}$. A directed path from $A$ to $B$ in such a graph indicates a *causal relation* $A \Rightarrow B$ in the system, where cause $A$ *influences* the value of its effect $B$, but not the other way around. An edge $A \rightarrow B$ in $\mathcal{G}_C$ indicates a *direct* causal link. In data from a system with a causal relation $A \Rightarrow B$ (direct or indirect), the values of $A$ and $B$ have a tendency to vary together, i.e. they become probabilistically dependent.

Two assumptions are usually employed to link the underlying, asymmetric causal relations to observable, symmetric probabilistic dependencies:

The **Causal Markov Condition** states that each variable in a causal DAG $\mathcal{G}_C$ is (probabilistically) independent of its non-descendants, given its parents.
The **Causal Faithfulness Condition** states that there are no independencies between variables that are not entailed by the Causal Markov Condition.

The first makes it possible to go from causal graph to observed probabilistic independencies; the second completes the way back. Together, they imply that the causal DAG $\mathcal{G}_C$ is also *minimal*, in the sense that no proper subgraph can satisfy both assumptions *and* produce the same probability distribution (Zhang and Spirtes, 2008).

Before we continue with causal discovery methods, first a quick recap of some standard concepts and terminology. The joint probability distribution induced by a causal DAG $\mathcal{G}_C$ factors according to a *Bayesian network* (BN): a pair $\mathcal{B} = (\mathcal{G}, \boldsymbol{\Theta})$, where $\mathcal{G} = (\mathbf{V}, \mathbf{A})$ is DAG over random variables $\mathbf{V}$, and the parameters $\boldsymbol{\theta}_V \subset \boldsymbol{\Theta}$ represent the conditional probability of variable $V \in \mathbf{V}$ given its parents $\mathbf{Pa}(V)$ in the graph $\mathcal{G}$. Probabilistic independencies can be read from the graph $\mathcal{G}$ via the well-known *d-separation* criterion: $X$ is conditionally independent of $Y$ given $\mathbf{Z}$, denoted $X \perp\!\!\!\perp Y \mid \mathbf{Z}$, iff there is no unblocked path between $X$ and $Y$ in $\mathcal{G}$ conditional on the nodes in $\mathbf{Z}$; see e.g. (Pearl, 1988; Neapolitan, 2004).

If some of the variables in the causal DAG are hidden then the independence relations between the observed variables may be represented in the form of a *(maximal) ancestral graph* (MAG) (Richardson and Spirtes, 2002). MAGs form an extension of the class of DAGs that is closed under marginalization and selection. In addition to directed arcs, MAGs can also contain *bidirected arcs* $X \leftrightarrow Y$ (indicative of marginalization) and *undirected edges* $X \text{---} Y$ (indicative of selection). The *causal sufficiency* assumption states that there are no hidden common causes of the observed variables in $\mathcal{G}$, which implies that the distribution over the observed variables still conforms to a Bayesian network. In this article we ignore selection bias (= no undirected edges), but do *not* rely on causal sufficiency.

The *equivalence class* $[\mathcal{G}]$ of a graph $\mathcal{G}$ is the set of all graphs that are indistinguishable in terms of (Markov) implied independencies.[1] For a DAG or MAG $\mathcal{G}$, the corresponding equivalence class $[\mathcal{G}]$ can be represented as a *partial ancestral graph* (PAG) $\mathcal{P}$, which keeps the skeleton (adjacencies) and all invariant edge marks, i.e. tails (−) and arrowheads (>) that appear in all members of the equivalence class, and turns the remaining non-invariant edge marks into circles (◦) (Zhang, 2008). The invariant arrowhead at an edge $A \mathbin{*\!\!\rightarrow} B$ in $\mathcal{P}$ signifies that $B$ is *not* a cause of $A$. An edge $A \rightarrow B$ implies a causal link $A \Rightarrow B$ that is also direct.

With this in mind, the task of a causal discovery algorithm is to find as many invariant features of the equivalence class corresponding to a given data set as possible. From this, all identifiable, present or absent causal relations can be read.

**Causal discovery procedures**

A large class of **constraint-based** causal discovery algorithms is based directly on the faithfulness assumption: if a conditional independence $X \perp\!\!\!\perp Y \mid \mathbf{Z}$ can be found for *any* set of variables $\mathbf{Z}$, then there is no direct causal relation between $X$ and $Y$ in the underlying causal graph $\mathcal{G}_C$, and hence no edge between $X$ and $Y$ in the equivalence class $\mathcal{P}$. In this way, an exhaustive search over all pairs of variables can uncover the entire skeleton of $\mathcal{P}$. In the subsequent stage a number of orientation rules are executed that find the invariant tails and arrowheads.

Members of this group include the IC-algorithm (Pearl and Verma, 1991), PC/FCI (Spirtes et al., 2000), Grow-Shrink (Margaritis and Thrun, 1999), TC (Pellet and Elisseef, 2008), and many others. All involve repeated independence tests in the adjacency search phase, and employ orientation rules as described in Meek (1995). The differences lie mainly in the search strategy employed, size of the conditioning sets, and additional assumptions imposed. Of these, the FCI algorithm in conjunction with the additional orientation rules in (Zhang, 2008) is the only one that is sound and complete in the large-sample limit when hidden common causes and/or selection bias may be present.

Constraint-based procedures tend to output a single, reasonably clear graph, representing the class of all possible causal DAGs. The downside is that for finite data they give little indication of which parts of the network are stable (reliable), and which are not: if unchecked, even one erroneous, borderline independence decision may be propagated through the network, leading to multiple incorrect orientations (Spirtes, 2010).

To tackle the perceived lack of robustness of PC, Ramsey et al. (2006) proposed a **conservative approach** for the orientation phase. The standard rules draw on the implicit assumption that, after the initial adjacency search, a single $X \perp\!\!\!\perp Y \mid \mathbf{Z}$ should suffice to orient an unshielded triple $\langle X, Z, Y \rangle$, as $Z$ should be either part of *all* or part of *no* sets that separate $X$ and $Y$. The Conservative PC (CPC) algorithm tests explicitly whether this assumption holds, and only orients the triple into a noncollider resp. v-structure $X \rightarrow Z \leftarrow Y$ if found true. If not, then it is marked as *unfaithful*. Tests show that CPC significantly outperforms standard PC in terms of overall accuracy, albeit often with less informative output, for only a marginal increase in run-time.

This idea can be extended to FCI: the set of potential separating nodes is now conform FCI's adjacency search, and any of Zhang's orientation rules that relies on a particular unshielded (non-)collider does not fire on an unfaithful triple. See (Glymour et al., 2004; Kalisch et al., 2011) for an implementation of Conservative FCI (CFCI) and many related algorithms.

The **score-based** approach is an alternative paradigm that builds on the implied *minimality* of the causal graph: define a scoring criterion $S(\mathcal{G}, \mathbf{D})$ that measures how well a Bayesian network with structure $\mathcal{G}$ fits the observed data $\mathbf{D}$, while preferring simpler networks, with fewer free parameters, over more complex ones. If the causal relations between the variables in $\mathbf{D}$ form a causal DAG $\mathcal{G}_C$, then in the large sample limit the highest scoring structure $\mathcal{G}$ must be part of the equivalence class of $[\mathcal{G}_C]$.

An example is the (Bayesian) likelihood score: given a Bayesian network $\mathcal{B} = (\mathcal{G}, \boldsymbol{\Theta})$, the likelihood of observing a particular data set $\mathbf{D}$ can be computed recursively from the network. Integrating out the parameters $\boldsymbol{\Theta}$ in the conditional probability tables (CPTs) then results in:

$$p(\mathbf{D}|\mathcal{G}) = \int_{\boldsymbol{\Theta}} p(\mathbf{D}|\mathcal{G}, \boldsymbol{\Theta}) f(\boldsymbol{\Theta}|\mathcal{G}) \, d\boldsymbol{\Theta}, \qquad (1)$$

where $f$ is a conditional probability density function over the parameters $\boldsymbol{\Theta}$ given structure $\mathcal{G}$.

---

[1] We follow the standard assumption that Markov equivalence implies statistical equivalence (Spirtes, 2010).

A closed form solution to eq.(1) is used in algorithms such as K2 (Cooper and Herskovits, 1992) and the Greedy Equivalence Search (GES) (Chickering, 2002) to find an optimal structure by repeatedly comparing scores for slightly modified alternatives until no more improvement can be found. See also Bouckaert (1995) for an evaluation of different strategies using these and other measures such as the BIC-score and minimum description length.

Score-based procedures can output a set of high-scoring alternatives. This ambiguity makes the result arguably less straightforward to read, but does allow for a measured interpretation of the reliability of inferred causal relations, and is not susceptible to incorrect categorical decisions (Heckerman et al., 1999). The main drawback is the need to rely on the causal sufficiency assumption.

## 2 The Best of Both Worlds

The strength of a constraint-based algorithm like FCI is its ability to handle data from arbitrary faithful underlying causal DAGs and turn it into sound and clear, unambiguous causal output. The strength of the Bayesian score-based approach lies in the robustness and implicit confidence measure that a likelihood weighted combination of multiple models can bring.

▶ Our idea is to improve on conservative FCI by using a Bayesian approach to estimate the reliability of different constraints, and use this to decide if, when, and how that information should be used.

Instead of classifying pieces of information as reliable or not, we want to rank and process constraints according to a confidence measure. This should allow to avoid propagating unreliable decisions while retaining more confident ones. It also provides a principled means for conflict resolution. The end-result is hopefully a more informative output model than CFCI, while obtaining a higher accuracy than standard FCI can deliver.

To obtain a confidence measure that can be compared across different estimates we want to compute the *probability* that a given independence statement holds from a given data set $\mathbf{D}$. In an ideal Bayesian approach we could compute a likelihood $p(\mathbf{D}|\mathcal{M})$ for each $\mathcal{M} \in \mathbf{M}$ (see section 3 on how to approximate this). If we know that the set $\mathbf{M}$ contains the 'true' structure, then the probability of an independence hypothesis $I$ follows from normalized summation as:

$$p(I|\mathbf{D}) \propto \sum_{\mathcal{M} \in \mathbf{M}(I)} p(\mathbf{D}|\mathcal{M})p(\mathcal{M}), \qquad (2)$$

(Heckerman et al., 1999), where $\mathbf{M}(I)$ denotes the subset of structures that entail independence statement $I$, and $p(\mathcal{M})$ represents a prior distribution over the structures (see §3.4).

Two remarks. Firstly, it is well known that the number of possible graphs grows very quickly with the number of nodes $\mathbf{V}$. But eq.(2) equally applies when we limit data and structures to *subsets* of variables $\mathbf{X} \subset \mathbf{V}$. For sparse graphs we can choose to consider only subsets of size $K \ll |\mathbf{V}|$. We opt to go one step further and follow a search strategy similar to PC/FCI, using structures of increasing size. Secondly, it would be very inefficient to compute eq.(2) for each independence statement we want to evaluate. From a single likelihood distribution over structures over $\mathbf{X}$ we can immediately compute the probability of *all* possible independence statements between variables in $\mathbf{X}$, including complex combinations such as those implied by $v$-structures, just by summing the appropriate contributions for each statement.

Having obtained probability estimates for a list of in/dependence statements $\mathcal{I}$, we can rank these in decreasing order of reliability, and keep the ones based on a decision threshold $p(I|\mathbf{D}) > \theta$, with $\theta = 0.5$ as intuitive default. In case of remaining conflicting statements, the ones with higher confidence take precedence. The resulting algorithm is outlined below:

---
**Algorithm 1** Outline
    **Start** : database $\mathbf{D}$ over variables $\mathbf{V}$
    *Stage 1 - Adjacency search*
1:   fully connected graph $\mathcal{P}$, empty list $\mathcal{I}$, $K = 0$
2:   **repeat**
3:     **for all** $X - Y$ still connected in $\mathcal{P}$ **do**
4:       **for all** adjacent sets $\mathbf{Z}$ of $K$ nodes in $\mathcal{P}$ **do**
5:         estimate $p(\mathcal{M}|\mathbf{D})$ over $\{X, Y, \mathbf{Z}\}$
6:         sum to $p(I|\mathbf{D})$ for independencies $I$
7:         update $\mathcal{I}$ and $\mathcal{P}$ for each $p(I|\mathbf{D}) > \theta$
8:       **end for**
9:     **end for**
10:    $K = K + 1$
11: **until** all relevant found
    *Stage 2 - Orientation rules*
12: rank and filter $\mathcal{I}$ in decreasing order of reliability
13: orient unshielded triples in $\mathcal{P}$
14: run remaining orientation rules
15: return causal model $\mathcal{P}$

---

In this form, the Bayesian estimates are only used to guide the adjacency search (update skeleton $\mathcal{G}$, l.7), and to filter the list of independencies $\mathcal{I}$ (l.12). Ideally, we would like the probabilities to guide the orientation phase as well. This implies processing the independence statements sequentially, in decreasing order of reliability. For that we can use a recently developed variant of the FCI algorithm (Claassen and Heskes, 2011), that breaks up the inference into a series of *modular* steps that can be executed in arbitrary order. It works by translating observed independence constraints into **logical statements** about the presence or absence of certain causal relations.

Using square brackets to indicate *minimal* sets of nodes: if a variable $Z$ either *makes* or *breaks* an independence relation between $\{X, Y\}$, then

1. $X \perp\!\!\!\perp Y \mid [\mathbf{W} \cup Z] \vdash (Z \Rightarrow X) \vee (Z \Rightarrow Y)$,

2. $X \not\!\perp\!\!\!\perp Y \mid \mathbf{W} \cup [Z] \vdash Z \not\Rightarrow (\{X,Y\} \cup \mathbf{W})$.

In words: from a minimal independence we infer the *presence* of at least one from two causal relations, whereas a dependence identifies the *absence* of causal relations. Subsequent causal statements follow from deduction on the properties *transitivity* $(X \Rightarrow Y) + (Y \Rightarrow Z) \vdash (X \Rightarrow Z)$, and *irreflexivity* $(X \not\Rightarrow X)$.

## 3 Sequential Causal Inference

This section discusses the steps needed to turn the previous idea into a working algorithm in the next section. Main issues are: probability estimates for logical causal statements from substructures, Bayesian likelihood computation, and inference from unfaithful DAGs. Proofs are detailed in (Claassen and Heskes, 2012).

A word on notation: we use $\mathbf{D}$ to denote a data set over variables $\mathbf{V}$ from a distribution that is faithful to some (larger) causal DAG $\mathcal{G}_C$. L denotes the set of logical causal statements $L$ over two or three variables in $\mathbf{V}$, of the form given in the r.h.s. of rules 1 and 2, above. We use $\mathbf{M_X}$ to represent the set of MAGs over $\mathbf{X}$, and $\mathbf{M_X}(L)$ to denote the subset that entails logical statement $L$. We also use $\mathcal{G}$ to explicitly indicate a DAG, $\mathcal{M}$ for a MAG, and $\mathcal{P}$ for a PAG.

### 3.1 A Modular Approach

In order to process available information in (decreasing) order of reliability we need to obtain probability estimates for logical statements on causal relations from data. Similar to eq.(2), this follows from summing the normalized posterior likelihoods of all MAGs that entail that statement through $m$-separation:

**Lemma 1.** The probability of a logical causal statement $L$ given a data set $\mathbf{D}$ is given by

$$p(L|\mathbf{D}) = \frac{\sum_{\mathcal{M} \in \mathbf{M}(L)} p(\mathbf{D}|\mathcal{M}) p(\mathcal{M})}{\sum_{\mathcal{M} \in \mathbf{M}} p(\mathbf{D}|\mathcal{M}) p(\mathcal{M})}, \quad (3)$$

using the notational conventions introduced above.

As stated, in many cases considering only a small subset of the variables in $\mathbf{V}$ is already sufficient to infer $L$. But that also implies that there are multiple subsets that imply $L$, each with different probability estimates. As these relate to different sets of variables, they should not be combined as in standard multiple hypothesis tests, but instead we want to look for the *maximum* value that can be found.

**Lemma 2.** Let $\mathbf{D}$ be a data set over variables $\mathbf{V}$. Then $\forall \mathbf{X} \subseteq \mathbf{V} : p(L|\mathbf{D}) \geq \sum_{\mathcal{M} \in \mathbf{M_X}(L)} p(\mathcal{M}|\mathbf{D})$.

*Proof.* Let $p(\mathcal{M}|\mathbf{D})$ be the posterior probability of MAG $\mathcal{M}$ given data $\mathbf{D}$. Let $\mathcal{M}(\mathbf{X})$ denote the MAG $\mathcal{M}$ marginalized to variables $\mathbf{X}$, then:

$$\begin{aligned} p(L|\mathbf{D}) &= \sum_{\mathcal{M} \in \mathbf{M_V}(L)} p(\mathcal{M}|\mathbf{D}) \\ &\geq \sum_{\mathcal{M} \in \mathbf{M_V}(L) : \mathcal{M}(\mathbf{X}) \in \mathbf{M_X}(L)} p(\mathcal{M}|\mathbf{D}) \\ &= \sum_{\mathcal{M} \in \mathbf{M_X}(L)} p(\mathcal{M}|\mathbf{D}) \end{aligned}$$

The inequality follows from the fact that, by definition, no marginal MAG $\mathcal{M}(\mathbf{X})$ entails a statement not entailed by $\mathcal{M}$, whereas the converse can (and does) occur. □

As a result: $p(L|\mathbf{D}) \geq \max_{\mathbf{X} \subset \mathbf{V}} \sum_{\mathcal{M} \in \mathbf{M_X}(L)} p(\mathcal{M}|\mathbf{D})$.

It means that while searching for logical causal statements L, it makes sense to keep track of the maximum probabilities obtained so far.

However, computing $p(\mathcal{M}|\mathbf{D})$ for $\mathcal{M} \in \mathbf{M_X}$ still involves computing likelihoods over all structures over $\mathbf{V}$, which is precisely what we want to avoid. A reasonable approximation is provided by $p(\mathcal{M}|\mathbf{D_X})$, i.e. the estimates obtained by only including data in $\mathbf{D}$ from the variables $\mathbf{X}$. It means that the lower bound is no longer guaranteed to hold universally, but should still be adequate in practice.

### 3.2 Obtaining likelihood estimates

If we know that the 'true' structure over a subset $\mathbf{X} \subseteq \mathbf{V}$ takes the form of a DAG, then computing the required likelihood estimates $p(\mathbf{D_X}|\mathcal{G})$ is relatively straightforward. Cooper and Herskovits (1992) showed that, under some reasonable assumptions, for discrete random variables the integral (1) has a closed-form solution. In the form presented in (Heckerman et al., 1995) this score is known as the *Bayesian Dirichlet* (BD) metric:

$$p(\mathbf{D}|\mathcal{G}) = \prod_{i=1}^{n} \prod_{j=1}^{q_i} \frac{\Gamma(N'_{ij})}{\Gamma(N_{ij} + N'_{ij})} \prod_{k=1}^{r_i} \frac{\Gamma(N_{ijk} + N'_{ijk})}{\Gamma(N'_{ijk})}, \quad (4)$$

with $n$ the number of variables, $r_i$ the multiplicity of variable $X_i$, $q_i$ the number of possible instantiations of the parents of $X_i$ in $\mathcal{G}$, $N_{ijk}$ the number of cases in data set $\mathbf{D}$ in which variable $X_i$ has the value $r_{i(k)}$ while its parents are instantiated as $q_{i(j)}$, and with $N_{ij} = \sum_{k=1}^{r_i} N_{ijk}$. The $N'_{ij} = \sum_{k=1}^{r_i} N'_{ijk}$ represent the pseudocounts for a Dirichlet prior over the parameters in the corresponding CPTs.

Different strategies for choosing the prior exist: for example, choosing $N'_{ijk} = 1$ (uniform prior) leads to the original *K2-metric*, see (Cooper and Herskovits, 1992). Setting $N'_{ijk} = N'/(r_i q_i)$ gives the popular *BDeu-metric*, which is *score equivalent* in the sense that structures from the same equivalence class $[\mathcal{G}]$ receive the same likelihood score, cf. (Buntine, 1991). In this article, we opt for the $K$2-metric, as it seems more appropriate in causal settings (Heckerman et al., 1995). But having to consider only one instance of every equivalence class may prove a decisive advantage of the BDe(u)-metric in future extensions.

However, eq.(4) only applies to DAGs. We know that, even when assuming causal sufficiency applies for the variables $\mathbf{V}$, considering arbitrary subsets size $|\mathbf{X}| \geq 4$ in general will require MAG representations to account for common causes that are not in $\mathbf{X}$. Extending the derivation of eq.(4) requires additional assumptions on multiplicity and number of the hidden variables, and turns the nice closed-form solution into an intractable problem that requires approximation, e.g. through sampling (Heckerman et al., 1999). This would make each step in our approach much more expensive. Recently, Evans and Richardson (2010) showed a maximum likelihood approach to fit acyclic directed mixed graphs (a superset of MAGs) directly on binary data. Unfortunately, this method cannot provide the likelihood estimates per model we need for our purposes. Silva and Ghahramani (2009) do present a Bayesian approach, but need to put additional constraints on the distribution in the form of (cumulative) Gaussian models.

In short, even though we would like to use MAGs to compute $p(\mathbf{D_X}|\mathcal{M})$ directly in eq.(3), at the moment we have to rely on DAGs to obtain approximations to the 'true' value. This will result in less accurate reliability estimates for $p(L|\mathbf{D})$, but also means that we may miss certain pieces of information, or, even worse, that the inference may become invalid.

### 3.3 Unfaithful inference: DAGs vs. MAGs

Fortunately we can show that, even when the true independence structure over a subset $\mathbf{X} \subset \mathbf{V}$ is a MAG, we can still do valid inference via $p(L|\mathbf{D_X})$ from likelihood scores over an exhaustive set of DAGs over $\mathbf{X}$, provided we account for unfaithful DAG representations. This part discusses unfaithful inference, and how the mapping from structures to logical causal statements can be modified. Much of what we show builds on (Bouckaert, 1995).

In the large-sample limit, the Bayesian likelihood score picks the smallest DAG structure(s) that can capture the observed probability distribution exactly.

**Definition 1** (**Optimal uDAG**). A DAG $\mathcal{G}$ is an (unfaithful) **uDAG** approximation to a MAG $\mathcal{M}$ over a set of nodes $\mathbf{X}$, iff for any probability distribution $p(\mathbf{X})$, generated by an underlying causal graph faithful to $\mathcal{M}$, there is a set of parameters $\boldsymbol{\Theta}$ such that the Bayesian network $\mathcal{B} = (\mathcal{G}, \boldsymbol{\Theta})$ encodes the same distribution $p(\mathbf{X})$. The uDAG is **optimal** if there exists no uDAG to $\mathcal{M}$ with fewer free parameters.

A uDAG is just a DAG for which we do not know if it is faithful or not. Reading in/dependence relations from a uDAG goes as follows:

**Lemma 3.** Let $\mathcal{B} = (\mathcal{G}, \boldsymbol{\Theta})$ be a Bayesian network over a set of nodes $\mathbf{X}$, with $\mathcal{G}$ a uDAG for some MAG that is faithful to a distribution $p(\mathbf{X})$. Let $\mathcal{G}_{X\|Y}$ be the graph obtained by eliminating the edge $X - Y$ from $\mathcal{G}$ (if present). Then, if $X \perp\!\!\!\perp_{\mathcal{G}_{X\|Y}} Y \mid \mathbf{Z}$ then:

$$(X \perp\!\!\!\perp_{\mathcal{G}} Y \mid \mathbf{Z}) \Leftrightarrow (X \perp\!\!\!\perp_p Y \mid \mathbf{Z}).$$

*Proof sketch.* The independence rule $(X \perp\!\!\!\perp_{\mathcal{G}} Y \mid \mathbf{Z}) \Rightarrow (X \perp\!\!\!\perp_p Y \mid \mathbf{Z})$ follows (Pearl, 1988). The dependence rule $(X \perp\!\!\!\perp_{\mathcal{G}_{X\|Y}} Y \mid \mathbf{Z}) \wedge (X \not\perp\!\!\!\perp_{\mathcal{G}} Y \mid \mathbf{Z}) \Rightarrow (X \not\perp\!\!\!\perp_p Y \mid \mathbf{Z})$ is similar to the 'coupling' theorem (3.11) in (Bouckaert, 1995), but stronger. As we assume a faithful MAG, a dependence $X \not\perp\!\!\!\perp_p Y \mid \mathbf{Z}$ cannot be destroyed by in/excluding a node $U$ that has no unblocked path in the underlying MAG to $X$ and/or $Y$ given $\mathbf{Z}$. This eliminates one of the preconditions in the coupling theorem. See (Claassen and Heskes, 2012) for details. □

So, in a uDAG all independencies from *d*-separation are still valid, but the identifiable dependencies are restricted.

**Example 1.** Treating the uDAG in Figure 1(b) as a faithful DAG would suggest $X \perp\!\!\!\perp_p T \mid [Z]$, and hence $(Z \Rightarrow X) \vee (Z \Rightarrow T)$. This is wrong: Figure 1(a) shows that $Z$ is ancestor of neither $X$ nor $T$. Lemma 3 would not make this mistake, as it allows to deduce $X \perp\!\!\!\perp_p T \mid Z$, but not the erroneous $X \not\perp\!\!\!\perp_p T$.

We can generalize Lemma 3 to *indirect* dependencies.

**Lemma 4.** Let $\mathcal{G}$ be a uDAG for a faithful MAG $\mathcal{M}$. Let $X$, $Y$, and $\mathbf{Z}$ be disjoint (sets of) nodes. If $\pi = \langle X, .., Y \rangle$ is the *only* unblocked path from $X$ to $Y$ given $\mathbf{Z}$ in $\mathcal{G}$, then $X \not\perp\!\!\!\perp_p Y \mid \mathbf{Z}$.

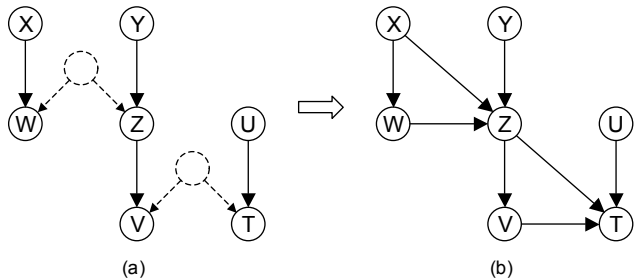

Figure 1: (a) causal DAG with hidden variables, (b) uDAG with unfaithful $X \perp\!\!\!\perp_{\mathcal{G}} T \mid [Z]$

**Example 2.** With Lemma 4 we infer from Figure 1(b) that $X \not\perp\!\!\!\perp_p T \mid \{V, W\}$. We also find that $Y \not\perp\!\!\!\perp_p V$, from which, in combination with $Y \perp\!\!\!\perp_p V \mid [Z]$, we (rightly) conclude that $(Z \Rightarrow Y) \vee (Z \Rightarrow V)$, see (a).

In general, Lemmas 3 and 4 assert different dependencies for different uDAG members of the same equivalence class. If the uDAG $\mathcal{G}$ is *optimal*, then all in/dependence statements from any uDAG member of the corresponding equivalence class $[\mathcal{G}]$ are valid. In that case we can do the inference based on the PAG representation $\mathcal{P}$ of $[\mathcal{G}]$. This provides additional information, but also simplifies some inference steps. Again, see (Claassen and Heskes, 2012) for details.

For example, identifying an absent causal relation (arrowhead) $X \not\Rightarrow Y$ from an optimal uDAG becomes identical to the inference from a faithful MAG. Let a *potentially directed path* (p.d.p.) be a path in a PAG that could be oriented into a directed path by changing circle marks into appropriate tails/arrowheads, then

**Lemma 5.** Let $\mathcal{G}$ be an optimal uDAG to a faithful MAG $\mathcal{M}$, then the absence of a causal relation $X \not\Rightarrow Y$ can be identified, iff there is no potentially directed path from $X$ to $Y$ in the PAG $\mathcal{P}$ of $[\mathcal{G}]$.

*Proof sketch.* The optimal uDAG $\mathcal{G}$ is obtained by (only) adding edges between variables in the MAG $\mathcal{M}$ to eliminate invariant bi-directed edges, until no more are left. At that point the uDAG is a representative of the corresponding equivalence class $\mathcal{P}$ (Theorem 2 in Zhang (2008)). For any faithful MAG all and only the nodes not connected by a p.d.p. in the corresponding PAG have a definite non-ancestor relation in the underlying causal graph. At least one uDAG instance in the equivalence class of an optimal uDAG over a given skeleton leaves the ancestral relations of the original MAG intact. Therefore, any remaining invariant arrowhead in the PAG $\mathcal{P}$ matches a non-ancestor relation in the original MAG. □

For the presence of causal relations (tails) a similar, but more complicated criterion can be found; see Supplement. Ultimately, the impact of having to use uDAGs boils down to a modified mapping of structures to logical causal statements, based on the inference rules above.

Finally, it is worth mentioning that in the large-sample limit, matching uDAGs over increasing sets of nodes we are guaranteed to find all independencies needed to obtain the skeleton, as well as all invariant arrowheads and many invariant tails. However, as the primary goal remains to improve accuracy/robustness when the large-sample limit does *not* apply, we do not pursue this matter further here.

### 3.4 Consistent prior over structures

The computation of $p(L|\mathbf{D_X})$ requires a prior distribution $p(\mathcal{M})$ over the set of MAGs over $\mathbf{X}$. A straightforward solution is to use a uniform prior, assigning equal probability to each $\mathcal{M} \in \mathbf{M}$. Alternatively, we can use a predefined function that penalizes complexity or deviation w.r.t. some reference structure (Chickering, 2002; Heckerman et al., 1995). If we want to exploit score-equivalence with the BDe(u) metric in eq.(4), we can weight DAG representatives according to the size of their equivalence class.

If we have background information on expected (or desired) properties of the structure, such as max. node degree, average connectivity, or small-world/scale-free networks, we can use this to construct a prior $p(\mathcal{M})$ through *sampling*: generate random graphs over all variables in accordance with the specified characteristics, sample one or more random subsets of variables size $K$, and compute the marginal structure over that subset. Averaging over structures that are PAG-isomorphs (equivalence classes identical under relabeling) improves both consistency and convergence.

Irrespective of the method, it is essential to ensure the prior is also *consistent* over structures of different size. Perhaps surprisingly, this is *not* obtained by applying the same strategy at different levels: a uniform distribution over DAGs over $\{X, Y, Z\}$ implies $p(``X \perp\!\!\!\perp Y") = 6/25$, whereas a uniform distribution over two-node DAGs implies $p(``X \perp\!\!\!\perp Y") = 1/3$. We obtain a consistent multi-level prior by starting from a preselected level $K$, and then extend to different sized structures through marginalization.

## 4 The BCCD algorithm

We can now turn the results from the previous section into a working algorithm. The implementation largely follows the outline in Algorithm 1, except that now uDAGs (instead of MAGs) are used to obtain a list of logical causal statements (instead of independencies), and that logical inference takes the place of the orientation rules, resulting in the Bayesian Constraint-based Causal Discovery (BCCD) algorithm.

A crucial step in the algorithm is the mapping $\mathcal{G} \times L$ from optimal uDAG structures to causal statements in line 9. This mapping is the same for each run, so it can be precomputed from the rules in section 3.3 and the Supplement, and stored for use afterwards (l.1) The uDAGs $\mathcal{G}$ are represented as adjacency matrices. For speed and efficiency purposes, we choose to limit the structures to size $K \leq 5$, which gives a list of 29,281 uDAGs at the highest level. For details about representation and rules, see (Claassen and Heskes, 2012).

---

**Algorithm 2** Bayesian Constr. Causal Discovery

    **In** : database **D** over variables **V**, backgr.info $\mathcal{I}$
    **Out**: causal relations matrix $\mathbf{M}_C$, causal PAG $\mathcal{P}$
    *Stage 0 - Mapping*
1:  $\mathcal{G} \times L \leftarrow Get\_uDAG\_Mapping(\mathbf{V}, K_{max} = 5)$
2:  $p(\mathcal{G}) \leftarrow Get\_Prior(\mathcal{I})$
    *Stage 1 - Search*
3:  fully connected $\mathcal{P}$, empty list L, $K = 0$, $\theta = 0.5$
4:  **while** $K \leq K_{max}$ **do**
5:     **for all** $X \in \mathbf{V}, Y \in Adj(X)$ in $\mathcal{P}$ **do**
6:         **for all** $\mathbf{Z} \subseteq Adj(X)_{\setminus Y}, |\mathbf{Z}| = K$ **do**
7:             $\mathbf{W} \leftarrow Check\_Unprocessed(X, Y, \mathbf{Z})$
8:             $\forall \mathcal{G} \in \mathcal{G}_\mathbf{W}$ : compute $p(\mathcal{G}|\mathbf{D_W})$
9:             $\forall L : p(L_\mathbf{W}|\mathbf{D_W}) \leftarrow \sum_{\mathcal{G} \rightarrow L_\mathbf{W}} p(\mathcal{G}|\mathbf{D_W})$
10:           $\forall L : p(L) \leftarrow \max(p(L), p(L_\mathbf{W}|\mathbf{D_W}))$
11:           $\mathcal{P} \leftarrow p(\text{"}W_i \not\asymp W_j\text{"}|\mathbf{D_W}) > \theta$
12:         **end for**
13:     **end for**
14:     $K = K + 1$
15: **end while**
    *Stage 2 - Inference*
16: $\mathbf{L}_C =$ empty 3D-matrix size $|\mathbf{V}|^3$, $i = 1$
17: $L \leftarrow Sort\_Descending(L, p(L))$
18: **while** $p(L_i) > \theta$ **do**
19:     $\mathbf{L}_C \leftarrow Run\_Causal\_Logic(\mathbf{L}_C, L_i)$
20:     $i \leftarrow i + 1$
21: **end while**
22: $\mathbf{M}_C \leftarrow Get\_Causal\_Matrix(\mathbf{L}_C)$
23: $\mathcal{P} \leftarrow Map\_To\_PAG(\mathcal{P}, \mathbf{M}_C)$

---

We verified the resulting mapping from uDAG rules to logical statements against a *brute-force approach* that checks the intersection of all logical causal statements implied by all MAGs that can have a particular uDAG as an optimal approximation. We found that at least for uDAG structures up to five nodes our rules are still sound and complete. Interestingly enough, for uDAGs up to four nodes all implied statements are *identical* to those that would be obtained if we just treated uDAGs as faithful DAGs. Only at five+ nodes do we have to take into account that uDAGs are unfaithful.

The adjacency search (l.3-15), loops over subsets from neigbouring nodes for identifiable causal information, while keeping track of adjacencies that can be eliminated (l.11). For structures over five or more nodes we need to consider nodes from FCI's *Possible-D-Sep* set (Spirtes et al., 1999). In practice, it rarely finds any additional (reliable) independencies, and we opt to skip this step for speed and simplicity (l.6), similar to the RFCI approach in (Colombo et al., 2011). As the set $\mathbf{W} = \{X, Y\} \cup \mathbf{Z}$ can be encountered in different ways, line (7) checks if the test on that set has been performed already. A list of probability estimates $p(L|\mathbf{D})$ for each logical causal statement is built up (l.10), until no more information is found.

The inference stage (l.16-21) then processes the list L in decreasing order of reliability, until the threshold is reached. Statements in L are added one-by-one to the matrix of logical causal statements $\mathbf{L}_C$ (encoding identical to L), with additional information inferred from the causal logic rules. Basic conflict resolution is achieved through not overriding existing information (from more reliable statements). The final step (l.22,23) retrieves all explicit causal relations in the form of a causal matrix $\mathbf{M}_C$, and maps this onto the skeleton $\mathcal{P}$ obtained from Stage 1 to return a graphical PAG representation.

## 5 Experimental Evaluation

We have tested various aspects of the BCCD algorithm in many different circumstances, and against various other methods. The principal aim at this stage is to verify the viability of the Bayesian approach. We compare our results and that of other methods from data against known ground-truth causal models. For that, we generate random causal graphs with certain predefined properties (adapted from Melancon et al. (2000); Chung and Lu (2002)), generate random data from this model, and marginalize out one or more hidden confounders. We looked at the impact of the number of data points, size of the models, sparseness, choices for parameter settings etc. on the performance to get a good feel for expected strengths and weaknesses in real-world situations.

It is well-known that the relative performance of different causal discovery methods can depend strongly on the performance metric and/or specific test problems used in the evaluation. Therefore, we will not claim that our method is inherently better than others based on the experimental results below, but simply note that the fact that in nearly all test cases the BCCD algorithm performed as good or better than other methods, is a clear indication of the viability and potential of this approach.

Having limited space available, we only include results of tests against the two other state-of-the-art methods that can handle hidden confounders: FCI as the de facto benchmark, and its equivalent adapted from conservative PC. For the evaluation we use two complementary metrics: the *PAG accuracy* looks at the graphical causal model output and counts the number of edge marks that matches the PAG of true equivalence class (excluding self-references). The *causal accuracy* looks at the proportion of all causal decisions, either explicit as BCCD does or implicit from the PAG for FCI, that are correct compared to the generating causal graph.

In a nutshell: we found that in most circumstances conservative FCI outperforms vanilla FCI by about $3-4\%$ in terms of PAG accuracy and slightly more in terms of causal accuracy. In its standard form, with a uniform prior over structures of 5 nodes, the BCCD algorithm consistently outperforms conservative FCI by a small margin of about $1-2\%$ at default decision thresholds ($\theta = 0.5$ for BCCD, $\alpha = 0.05$ for FCI). Including additional tests / nodes per test and using an extended mapping often increases this difference to about $2-4\%$ at optimal settings for both approaches (cf. Figure 3). This gain does come at a cost: BCCD has an increase in run-time of about a factor two compared to conservative FCI, which in turn is marginally more expensive than standard FCI. Evaluating many large structures can increases this cost even further, unless we switch to evaluating equivalence classes via the BDe metric in §3.2. However, the main benefit of the BCCD approach lies not in a slight improvement in accuracy, but in the added insight it provides into the generated causal model: even in this simple form, the algorithm gives a useful indication of which causal decisions are reliable and which are not, which seems very useful to have in practice.

The figures below illustrate some of these findings.

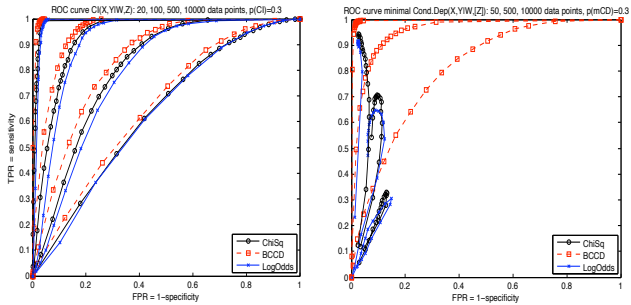

Figure 2: BCCD approach to (complex) independence test in eq.(2); (a) conditional independence $X \perp\!\!\!\perp Y \mid W, Z$, (b) *minimal* conditional dependence $X \not\!\perp\!\!\!\perp Y \mid W \cup [Z]$

First we implemented the BCCD approach as a (minimal) independence test. Figure 2 shows a typical example in the form of ROC-curves for different sized data sets, compared against a chi-squared test and a Bayesian log-odds test from (Margaritis and Bromberg, 2009), with the prior on independence as the tuning parameter for BCCD. For 'regular' conditional independence there was no significant difference (BCCD slightly ahead), but other methods reject minimal independencies for both high and low decision thresholds, resulting in the looped curves in (b); BCCD has no such problem.

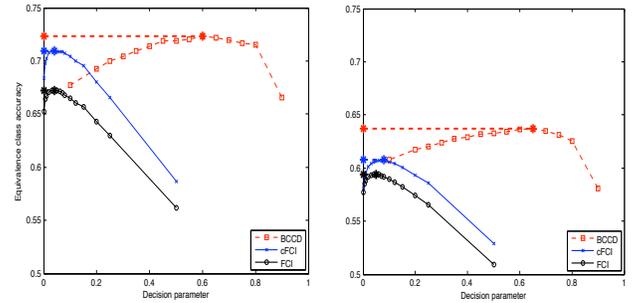

Figure 3: Equivalence class accuracy (% of edge marks in PAG) vs. decision parameter; for BCCD and (conservative) FCI, from 1000 random models; (a) 6 observed nodes, 1-2 hidden, 1000 points, (b) idem, 12 observed nodes

Figure 3 shows typical results for the BCCD algorithm itself: for a data set of 1000 records the *PAG accuracy* for both FCI and conservative FCI peaks around a threshold $\alpha \approx 0.05$ - lower for more records, higher for less - with conservative FCI consistently outperforming standard FCI. The BCCD algorithm peaks at a cut-off value $\theta \in [0.4, 0.8]$ with an accuracy that is slightly higher than the maximum for conservative FCI. The PAG accuracy tends not to vary much over this interval, making the default choice $\theta = 0.5$ fairly safe, even though the number of invariant edge marks does increase significantly (more decisions).

Table 5 shows the *confusion matrices* for edge marks in the PAG model at standard threshold settings for each of the three methods. We can recognize how FCI makes more explicit decisions than the other two (less circle marks), but also includes more mistakes. Conservative FCI starts from the same skeleton: it is more reluctant to orient uncertain $v$-structures, but manages to increase the overall accuracy as a result (sum of diagonal entries). The BCCD algorithm provides the expected compromise: more decisions than CFCI, but less mistakes than FCI, resulting in a modest improvement on the output PAG.

Figure 4 depicts the *causal accuracy* as a function of the tuning parameter for the three methods. The BCCD dependency is set against $(1-\theta)$ so that going

| BCCD | ✕ | → | — | —o |
|---|---|---|---|---|
| ✕ | 13.2 | 0.1 | 0.0 | 0.1 |
| → | 1.0 | 3.3 | 0.1 | 0.7 |
| — | 0.3 | 0.5 | 0.7 | 0.5 |
| —o | 1.6 | 1.7 | 0.8 | 5.5 |
|  | 16.1 | 5.6 | 1.6 | 6.8 |

(a)

| cFCI | ✕ | → | — | —o |
|---|---|---|---|---|
| ✕ | 12.7 | 0.3 | 0.0 | 0.4 |
| → | 0.9 | 3.2 | 0.0 | 0.9 |
| — | 0.3 | 0.4 | 0.4 | 0.9 |
| —o | 1.4 | 1.8 | 0.3 | 6.0 |
|  | 15.3 | 5.7 | 0.7 | 8.2 |

(b)

| FCI | ✕ | → | — | —o |
|---|---|---|---|---|
| ✕ | 12.7 | 0.5 | 0.0 | 0.2 |
| → | 0.9 | 3.5 | 0.0 | 0.6 |
| — | 0.3 | 0.9 | 0.4 | 0.4 |
| —o | 1.4 | 3.7 | 0.5 | 3.9 |
|  | 15.3 | 8.6 | 0.9 | 5.1 |

(c)

Table 1: Confusion matrices for PAG output: (a) BCCD algorithm, (b) Conservative FCI, (c) Standard FCI ; rows = true value, columns = output edge mark (1000 random models over 6 nodes, 10,000 data points)

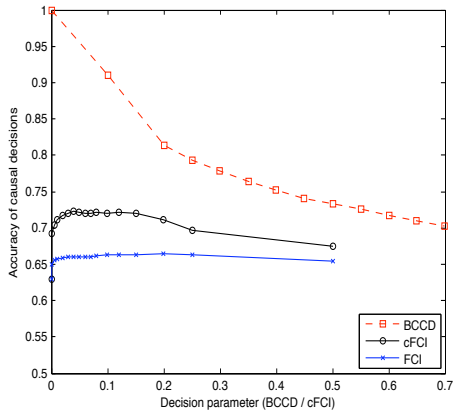

Figure 4: Accuracy of causal decisions as a function of the decision parameter

from $0 \to 1$ matches processing the list of statements in decreasing order of reliability. As hoped/expected: changing the decision parameter $\theta$ allows to access a range of accuracies, from a few very reliable causal relations to more but less certain indications. In contrast, the accuracy of the two FCI algorithms cannot be tuned effectively through the decision parameter $\alpha$. The reason behind this is apparent from Figure 2(b): changing the decision threshold in an independence test shifts the balance between dependence and independence decisions, but it cannot identify or alter the balance in favor of more reliable decisions. We consider the fact that the BCCD *can* do exactly that as the most promising aspect of the Bayesian approach.

## 6 Extensions and Future Work

The experimental results confirm that the Bayesian approach is both viable and promising: even in a basic implementation the BCCD algorithm already outperforms other state-of-the-art causal discovery algorithms. It yields slightly better accuracy, both for optimal and standard settings of the decision parameters. Furthermore, BCCD comes with a decision threshold that is easy to interpret and can be used to vary from making just a few but very reliable causal statements to many possibly less certain decisions. Perhaps counterintuitively, changing the confidence level in (conservative) FCI does not lead to similar behavior as it only affects the balance between dependence and independence decisions, which in itself does not increase the reliability of either.

An interesting question is how far off from the theoretical optimum we are: at the moment it is not clear whether we are fighting for the last few percent or if sizeable gains can still be made. There are many opportunities left for improvement, both in speed and accuracy. An easy option is to try to squeeze out as much as possible from the current framework: scoring equivalence classes with the BDe metric should bring a significant performance gain for large structures, without any obvious drawback, as the likelihood contributions of all members are aggregated anyway.

A hopeful but ambitious path is to tackle some of the fundamental problems: as stated, we would like to score MAGs directly instead of having to go via uDAGs. This would improve both the mapping and the probability estimates of the inferred logical causal statements. We can try to include higher order independencies (larger substructures) through sampling: reasonable probability estimate can be obained from a limited number of high scoring alternatives, see, e.g. (Bouckaert, 1995). Finally, we would like to obtain principled probability estimates for new statements derived during the inference process: this would improve conflict resolution, and would ultimately allow to give a meaningful estimate for the probability of all causal relations inferred from a given data set.

### Acknowledgements

This research was supported by NWO Vici grant nr.639.023.604.